%% file: paper.tex
\documentclass[sigconf]{acmart}

\usepackage[compact]{titlesec}
\titlespacing{\section}{0pt}{*0}{*0}
\titlespacing{\subsection}{0pt}{*0}{*0}

\usepackage{mathtools}

\def\BibTeX{{\rm B\kern-.05em{\sc i\kern-.025em b}\kern-.08emT\kern-.1667em\lower.7ex\hbox{E}\kern-.125emX}}

\usepackage{afterpage}
\usepackage{nicefrac}
\usepackage{xcolor}
\usepackage{siunitx}
\usepackage{array,framed}
\usepackage{booktabs}
\usepackage{
  color,
  float,
  epsfig,
  wrapfig,
  graphics,
  graphicx,
  subcaption
}
\usepackage{textcomp} 
\usepackage{setspace}
\usepackage{latexsym,fancyhdr,url}
\usepackage{enumerate}
\usepackage{algorithm2e}
\usepackage{algpseudocode}
\usepackage{graphics}
\usepackage{rotating}
\usepackage{xparse} 
\usepackage{xspace}
\usepackage{multirow}
\usepackage{csvsimple}
\usepackage{balance}
\usepackage{tabularx, booktabs}
\usepackage{comment}
\usepackage{soul}
\usepackage[skip=3pt,size=footnotesize]{caption}

\usepackage{
  tikz,
  pgfplots,
  pgfplotstable
}
\usepackage{hyperref}

\usetikzlibrary{
  shapes.geometric,
  arrows,
  external,
  pgfplots.groupplots,
  matrix
}

\pgfplotsset{compat=1.9}

\usepackage{mathtools,}

\DeclareMathAlphabet{\mathcal}{OMS}{cmsy}{m}{n}

\DeclareGraphicsExtensions{%
    .png,.PNG,%
    .pdf,.PDF,%
    .jpg,.mps,.jpeg,.jbig2,.jb2,.JPG,.JPEG,.JBIG2,.JB2}

\input{defs}
\usepackage{booktabs} 

\copyrightyear{2022}
\acmYear{2022}
\setcopyright{rightsretained}
\acmConference[HotMobile '22]{The 23rd International Workshop on
Mobile Computing Systems and Applications}{March 9--10, 2022}{Tempe,
AZ, USA}
\acmBooktitle{The 23rd International Workshop on Mobile Computing
Systems and Applications (HotMobile '22), March 9--10, 2022, Tempe,
AZ, USA}
\acmDOI{10.1145/3508396.3512877}
\acmISBN{978-1-4503-9218-1/22/03}

\newcommand{\xref}[1]{\S\ref{#1}}

\begin{document}
\fancyhead{}
\title{Towards Battery-Free Machine Learning and Inference in Underwater Environments}



\author{Yuchen Zhao}
\authornote{Co-primary authors}
\affiliation{
\institution{Imperial College London}
\country{}
}
\email{yuchen.zhao19@imperial.ac.uk}

\author{Sayed Saad Afzal}
\authornotemark[1]
\affiliation{
\institution{MIT}
\country{}
}
\email{afzals@mit.edu}

\author{Waleed Akbar}
\affiliation{
\institution{MIT}
\country{}
}
\email{wakbar@mit.edu}

\author{Osvy Rodriguez}
\affiliation{
\institution{MIT}
\country{}
}
\email{osvyrd@mit.edu}

\author{Fan Mo}
\affiliation{
\institution{Imperial College London}
\country{}
}
\email{f.mo18@imperial.ac.uk}

\author{David Boyle}
\affiliation{
\institution{Imperial College London}
\country{}
}
\email{david.boyle@imperial.ac.uk}

\author{Fadel Adib}
\affiliation{
\institution{MIT}
\country{}
}
\email{fadel@mit.edu}

\author{Hamed Haddadi}
\affiliation{
\institution{Imperial College London}
\country{}
}
\email{h.haddadi@imperial.ac.uk}


\input{0_abstract}

 \maketitle

\input{1_intro}
\input{2_architecture}
\input{3_feasibility}
\input{4_related}

\input{5_discussions}

\bibliographystyle{ACM-Reference-Format}
\balance
\bibliography{paper}

\end{document}

%% file: defs.tex
\usepackage{xparse}
\newcommand{\bnm}{\begin{newmath}}
\newcommand{\enm}{\end{newmath}}

\newcommand{\bea}{\begin{eqnarray*}}%
\newcommand{\eea}{\end{eqnarray*}}%

\newcommand{\bne}{\begin{newequation}}
\newcommand{\ene}{\end{newequation}}

\newcommand{\bal}{\begin{newalign}}
\newcommand{\eal}{\end{newalign}}

\newenvironment{newalign}{\begin{align}%
\setlength{\abovedisplayskip}{4pt}%
\setlength{\belowdisplayskip}{4pt}%
\setlength{\abovedisplayshortskip}{6pt}%
\setlength{\belowdisplayshortskip}{6pt} }{\end{align}}

\newenvironment{newmath}{\begin{displaymath}%
\setlength{\abovedisplayskip}{4pt}%
\setlength{\belowdisplayskip}{4pt}%
\setlength{\abovedisplayshortskip}{6pt}%
\setlength{\belowdisplayshortskip}{6pt} }{\end{displaymath}}

\newenvironment{newequation}{\begin{equation}%
\setlength{\abovedisplayskip}{4pt}%
\setlength{\belowdisplayskip}{4pt}%
\setlength{\abovedisplayshortskip}{6pt}%
\setlength{\belowdisplayshortskip}{6pt} }{\end{equation}}

\newcounter{ctr}

\newcounter{mytable}
\def\mytable{\begin{centering}\refstepcounter{mytable}}
\def\endmytable{\end{centering}}

\newcounter{myfig}
\def\myfig{\begin{centering}\refstepcounter{myfig}}
\def\endmyfig{\end{centering}}

\newlength{\saveparindent}
\newlength{\saveparskip}
\newcommand{\E}{{\rm I\kern-.3em E}}

\renewcommand{\eqref}[1]{\mbox{Equation~(\ref{#1})}}

\def \part {part}

\renewcommand{\paragraph}[1]{\vspace*{6pt}\noindent\textbf{#1}\;}

\def \blackslug{\hbox{\hskip 1pt \vrule width 4pt height 8pt
    depth 1.5pt \hskip 1pt}}
\def \qed{\quad\blackslug\lower 8.5pt\null\par}

\newcounter{mynote}[section]

\newcommand\ignore[1]{}

\newcounter{rcnote}[section]

\newcounter{mrnote}[section]

\newcounter{fknote}[section]

\newcounter{anote}[section]

\DeclareMathSymbol{\mlq}{\mathord}{operators}{``}
\DeclareMathSymbol{\mrq}{\mathord}{operators}{`'}

\newcommand{\rhf}[2]{R_{f, \gamma}}

\DeclareDocumentCommand{\edist}{o o}{
  \ensuremath{
    \IfNoValueTF{#1}{{d}}{{\sf d}(#1,#2)}
  }
}

\newcommand{\olrk}[1]{\ifx\nursymbol#1\else\!\!\mskip4.5mu plus 0.5mu\left(\mskip0.5mu plus0.5mu #1\mskip1.5mu plus0.5mu \right)\fi}

\NewDocumentCommand{\indseq}{ O{1} O{r} }{{#1}\ldots {#2}}

\newcommand{\mysub}[1]{\vspace{2pt} \noindent \textbf{#1}}

\newcommand{\ie}{i.e.}
\newcommand{\eg}{e.g.}

%% file: 0_abstract.tex
\begin{abstract}
This paper is motivated by a simple question: Can we design and build battery-free devices capable of machine learning and inference in underwater environments? An affirmative answer to this question would have significant implications for a new generation of underwater sensing and monitoring applications for environmental monitoring, scientific exploration, and climate/weather prediction.

To answer this question, we explore the feasibility of bridging advances from the past decade in two fields: battery-free networking and low-power machine learning. Our exploration demonstrates that it is indeed possible to enable battery-free inference in underwater environments. We designed a device that can harvest energy from underwater sound, power up an ultra-low-power microcontroller and on-board sensor, perform local inference on sensed measurements using a lightweight Deep Neural Network, and communicate the inference result via backscatter to a receiver. We tested our prototype in an emulated marine bioacoustics application, demonstrating the potential to recognize underwater animal sounds without batteries. Through this exploration, we highlight the challenges and opportunities for making underwater battery-free inference and machine learning ubiquitous.
\end{abstract}

%% file: 1_intro.tex
\section{Introduction}
\label{sec:intro}

In recent years, advances in battery-free sensing have enabled a range of novel applications including localization and environmental sensing. These advances have enabled long-term use of wireless sensors without the need for power supplies or batteries, hence enabling a transition towards a pollution-free sensing infrastructure for a variety of environmental sensing and monitoring applications such as detecting pollution and monitoring biodiversity. These applications are not limited to smart cities and urban environments, but also extend to climate change monitoring and environmental sensing in uninhabited areas such as forests, mines, remote areas, and space~\cite{,abdelhamid2020self,yang2021minegps}. 

The vast majority of existing battery-free technologies have been designed for land applications and very few have been engineered for the ocean. Only recently have researchers started looking into battery-free sensors for underwater environments. However, existing underwater battery-free technologies stop at energy harvesting for powering up sensors and backscatter for communicating raw sensory data~\cite{jang2019underwater,ghaffarivardavagh2020underwater,guida2020underwater}. Meanwhile, research into energy-efficient edge-based machine-learning (ML) models have enabled lightweight models to be used for on-device ML inference for a variety of applications. There have been a number of advances in model reduction, compression, and layer-wise quantization for various objectives such as privacy, energy, and efficiency~\cite{DBLP:journals/corr/abs-2008-02397, aloufi2020privacy, 8962332}. A fundamental challenge that remains is to perform time-series data analysis and ML on underwater battery-free sensors. One benefit of such capabilities is that they would enable us to limit data transmission, hence increasing the energy budget on underwater devices and their operation longevity. This is particularly important for usages of underwater battery-free sensors in ambient and remote monitoring applications for the ocean, where communication bandwidth is narrow and the sources for energy harvesting are limited \cite{srujana2015multi,saeed2018energy}.

\looseness=-1

In this paper, we propose and present a new challenge to the community, investigating the feasibility of battery-free ML in underwater environments, where extremely lightweight and task-specific deep neural network (DNN) models are executed on dedicated and highly-efficient underwater sensor nodes. We further investigate the tradeoffs between lightweight on-device analytics requirements for an exemplar case, which is marine mammal recognition in the ocean. This use case has many applications including monitoring biodiversity, understanding marine animal migration patterns, and even supporting the discovery of new species~\cite{undiscoveredspecies}.



Our investigation focuses on two critical aspects of underwater battery-free ML, which are (1) the capability (\ie, model accuracy) of lightweight DNN models on the tasks to be conducted in underwater environments, and (2) the feasibility of hosting the DNN models on low-power underwater computational devices such as microcontrollers.
%
Specifically on the task of marine mammal recognition, we design an end-to-end pipeline (\ie, from recording sounds to classification results on an underwater battery-free device) and evaluate our prototype's accuracy and power consumption. Our preliminary results indicate that, for a task of classifying four marine mammals, lightweight DNN models can achieve decent accuracy. With the help of existing energy harvesting technologies and customized circuit design, it is possible to run lightweight ML on underwater battery-free devices.  These results have important implications for both the mobile and ML communities, and pose exciting new opportunities for ubiquitous underwater battery-free ML.


%% file: 2_architecture.tex
\section{Underwater Battery-free Architecture}
\label{sec:arch}

\begin{figure}[!t]
    \centering
    \includegraphics[width=\linewidth]{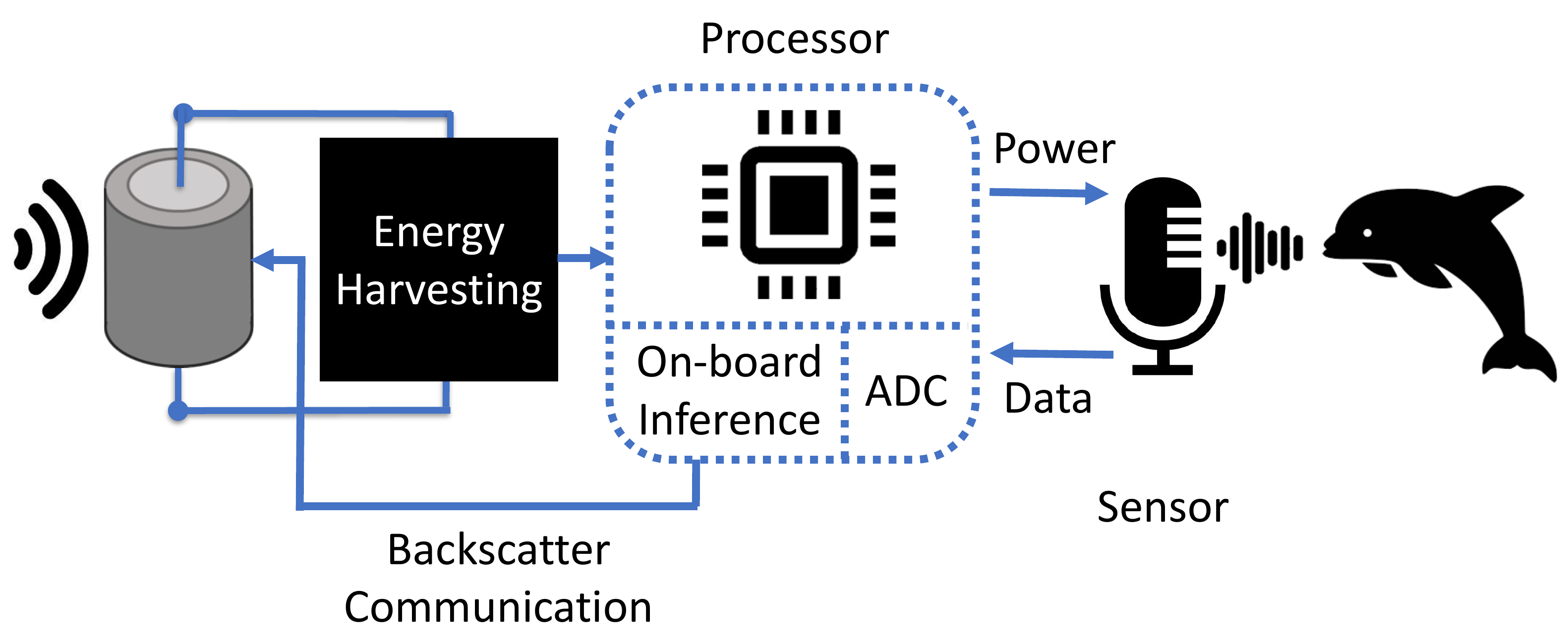}
    \caption{Design of battery-free inference on underwater edge nodes}
    \label{fig:system_design}
\end{figure}

We design a wireless, battery-free underwater system to perform sensing and inference on edge nodes and communicate with a receiver. To this end, we prototype a system that can perform marine mammal recognition in the ocean. Fig.~\ref{fig:system_design} shows the system design for our prototype. The system harvests energy from underwater sound to power up its processor and on-board sensor that captures animal sounds (\xref{sec:harvest}), performs on-board inference (\xref{sec:inference}), and transmits the result on the uplink via backscatter (\xref{sec:comms}). 

\begin{figure*}[t]
    \centering
    \includegraphics[width=\linewidth]{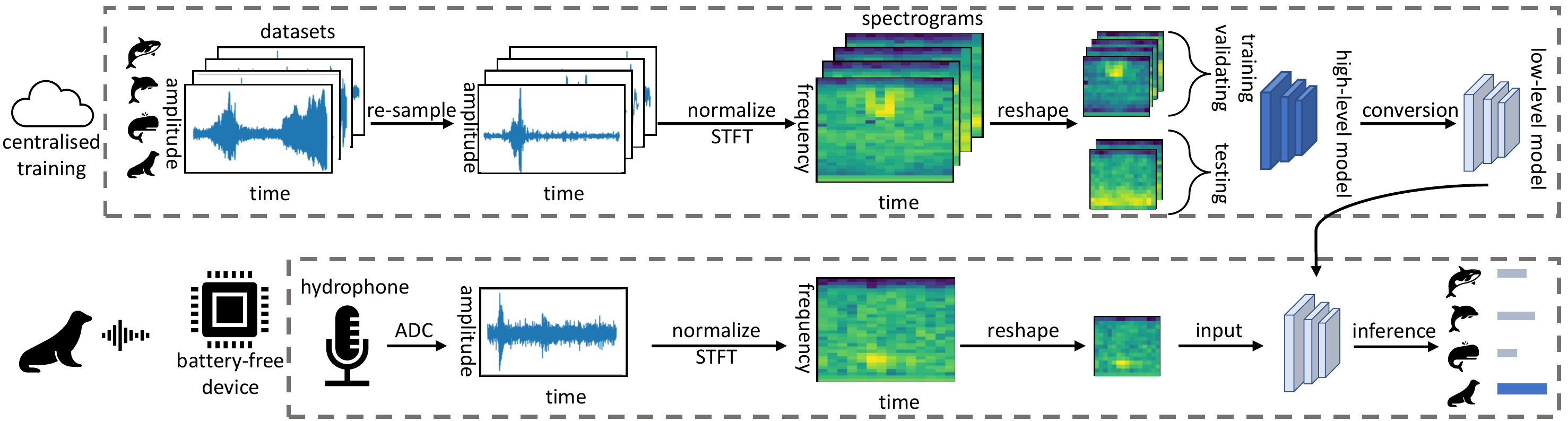}
    \caption{Overview of offline training and on-board inference. \textnormal{Initially, a high-level model is trained, validated, and evaluated offline on pre-collected audio clips from different marine mammals. A model with acceptable offline accuracy is converted to a low-level model that can be supported by standard C libraries on a target battery-free device. The converted model is then deployed in an end-to-end (\ie, from sounds received by a hydrophone to final classifications) marine mammal recognition pipeline on the target device.}}
    \label{fig:overview}
\end{figure*}

\subsection{Underwater Energy Harvesting}\label{sec:harvest}

To enable battery-free operation, our sensor needs to harvest energy from ambient underwater sources such as sound, waves, or temperature gradients. Recent past research has demonstrated that sound provides sufficient energy to power up low-power microcontrollers~\cite{jang2019underwater}; hence, we design our node to harvest energy from underwater sound. It is worth noting that our node architecture is general and can be adopted to other underwater ambient energy sources.

To harvest energy from sound, our node employs piezoelectric materials, which convert mechanical energy (i.e. sound) to electrical energy. Since the harvested signal is typically an alternating current (sound is a wave), our sensor node rectifies it to a DC voltage using a multi-stage rectifier. Once the node harvests sufficient energy, it can power up an on-board microcontroller. The microcontroller has an integrated ADC (analog to digital converter) which can be interfaced to a hydrophone (as shown in Fig.~\ref{fig:system_design}). The ADC samples received sound and stores it in memory.

\subsection{Battery-free Underwater Inference}\label{sec:inference}

The next stage in our design is to recognize animals from the recordings. To do this, one option is to program the node to transmit its recordings to a remote receiver that has a dedicated power source (e.g., an underwater drone or a coastal base station) and perform inference there. However, such an approach is undesirable for multiple reasons. First, since the throughput of underwater acoustic channels is limited, transmitting the full recording would require our node to remain powered up for an extended period of time, which will drain its harvested energy. Second, the limited throughput would incur a large delay in data transmission, which is detrimental in time-sensitive scenarios (e.g., detection of endangered species). Furthermore, if a desired receiver is not within the communication range of our node, the node would need to store the recording in its constrained memory, which limits its ability to process new recordings.  

To overcome these challenges, instead of transmitting the recordings for inference at a powered base station, our node performs on-board inference, as we describe in this section. This results in a lower-power, time-and-resource efficient, and more scalable  system design (as we show in~\xref{sec:feasibility}).

\subsubsection{Offline Training:}\label{offline_tr}

In order to have an accurate DNN model that can recognize marine mammals from their sounds, we need to use pre-collected audio clips (obtained from publicly available databases) to train the model until it achieves acceptable accuracy. Given the limited memory and computational resources of battery-free devices, it is difficult to perform the training process on them. Thus, we first train the high-level DNN models in a centralized manner (\eg, in the cloud) with offline training and validating data. 

To ensure that our model would perform well on the target underwater battery-free device, we introduce a number of pre-processing steps (\eg, re-sampling, normalization, and reshaping) as shown in Fig.~\ref{fig:overview}. These steps are necessary because the audio files (from databases) were collected using different equipment than our battery-free nodes (\eg, different ADC sampling rates and resolution). To mitigate the impact of these differences,
the raw data from pre-collected clips is re-sampled with the sample rate of the nodes.
The re-sampled data is then normalized and converted into a spectrogram through Short-Time Fourier transform (STFT).

To make the models work on a battery-free device, the high-level models need to be converted into a low-level model that satisfies the constraints of the architecture and memory of the target device. Our model conversion process uses techniques such as model compression and quantization to make the resulting low-level model work on the target device independently without support from run-time libraries.
\looseness=-1

\subsubsection{On-board Edge Inference:}

After the model is trained and converted, we can load the model parameters to the memory of the target device and then use it for on-board inference. We use the same pre-processing pipeline (normalization, STFT etc.) as for offline training which was discussed in~\xref{offline_tr}. One challenge with ultra-low-power microcontrollers is that, unlike standard audio cards, their ADCs cannot sample audio with both positive and negative amplitude. The lack of negative amplitude introduces another discrepancy between the pre-collected audio files and sound recordings collected using the microcontroller. To overcome this challenge, our design adds a clamping circuit to hardware design in order to passively add a DC offset to our input signal before we feed it to the on-board ADC. This DC offset is subsequently filtered in software by applying a high-pass filter; this ensures that the DC offset does not bias the STFT output or negatively impact the model performance.


\subsection{From Inference to Communication}\label{sec:comms}

The last step in our design is to communicate the inference result to a remote receiver. To do this, we leverage underwater backscatter -- a net-zero power communication technology -- to communicate  on the uplink, which allows us to reuse the same piezoelectric material for both harvesting (\xref{sec:harvest}) and communication. In underwater backscatter, the piezoelectric material is connected to an impedance switch that modulates its reflection coefficient. At a high level, by alternating between two impedance states, the node communicates in binary. A remote receiver can sense changes in reflected power and use them to decode the transmitted packet by applying standard underwater decoding mechanisms.
\looseness=-1



%% file: 3_feasibility.tex
\section{Feasibility Study}
\label{sec:feasibility}
We evaluated our prototype in terms of both its model accuracy and its power requirements. 


\subsection{Prototype Hardware}


\begin{figure}[t]
  \centering
\begin{tabular}{cc}
    \epsfig{file=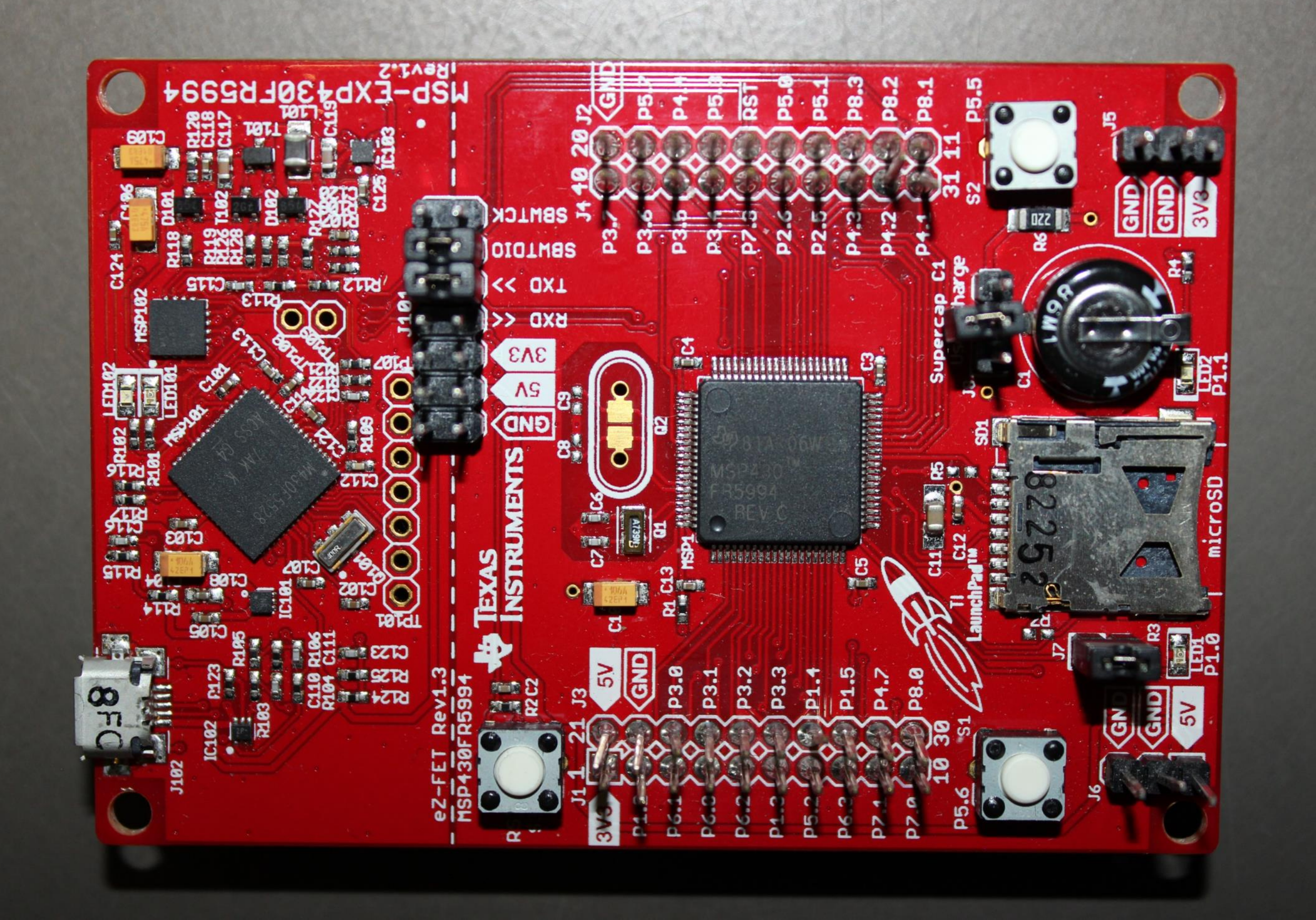,height=1.38in} & 
    \epsfig{file=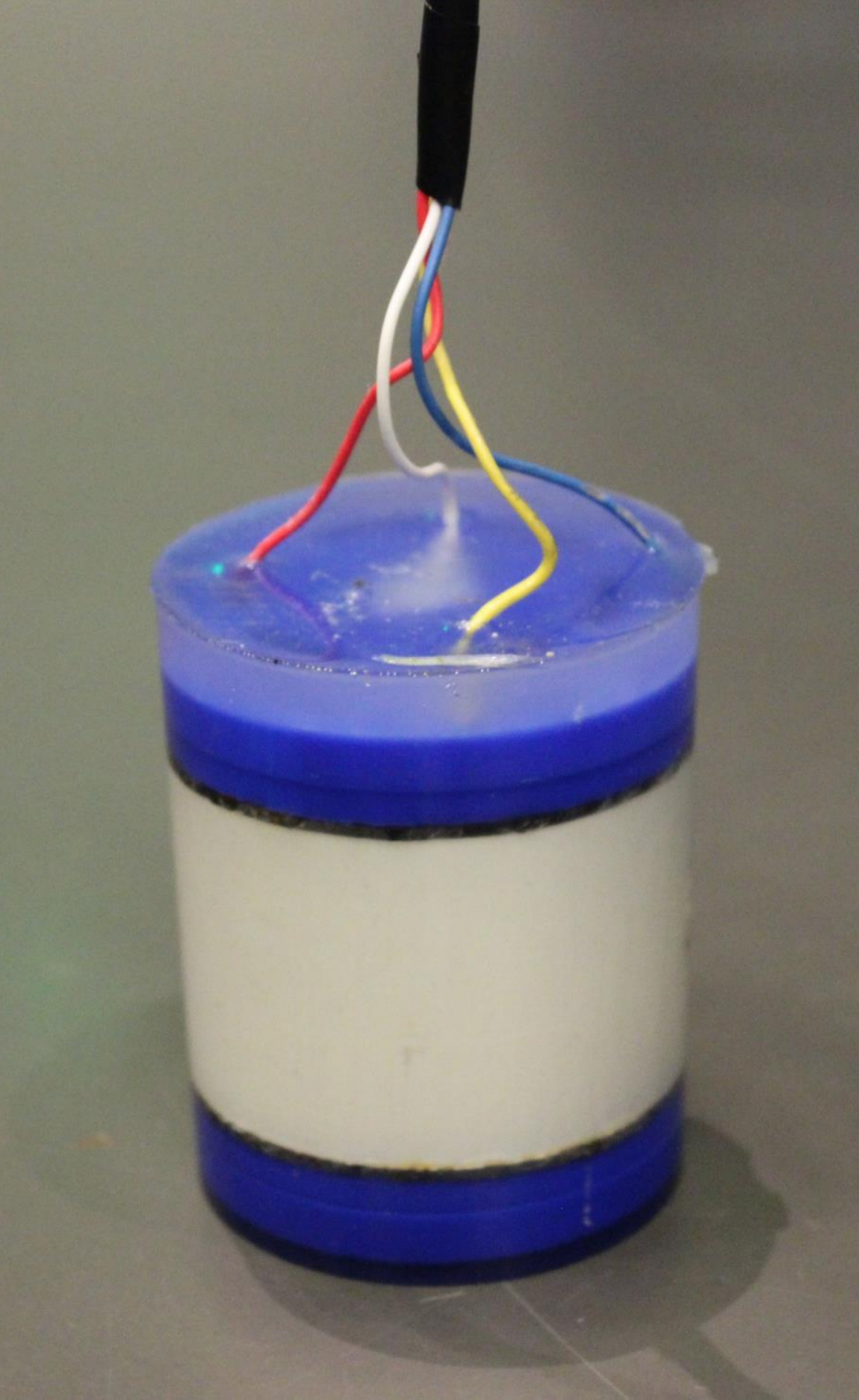,height=1.38in}
\\
\footnotesize (a) MSP430FR5994 MCU development board
&
\footnotesize (b) Potted Transducer
\end{tabular}
\caption{\textbf{Prototype Hardware} \textnormal{(a) shows the processor that our design uses for on-board inference. (b)
shows one of our potted transducers which can be used for energy harvesting and communication.}}
\label{fig:proto}
\end{figure}

\mysub{Backscatter node. }
Each backscatter node consists of a piezoelectric transducer and a hardware controller as shown in Fig.~\ref{fig:proto}. To build our transducers, we used piezo-ceramic cylinders with a nominal resonance frequency of 17 KHz ~\cite{steminc17}. Similar to our previous work, the nodes were potted, encapsulated for insulation and matching to water, and housed in 3D-printed mold and end-caps~\cite{saadafzal24}.

\mysub{Hardware controller.} The node hardware is used for energy harvesting, processing/inference, and interfacing with a sensor (e.g., a hydrophone). Our harvesting architecture consists of a standard multi-stage rectifier followed by a super-capacitor, and low dropout voltage regulator which drives the   digital logic unit. We implemented the logic on a MSP430-FR5994  microcontroller~\cite{MSP}. The microcontroller has a 12bit-ADC, which samples the (sound) signal at the rate of 330 samples/sec and stores a window of 512 samples in its SRAM. The microcontroller logic, including the code for downlink and uplink decoding as well as the inference model, is all stored in the FRAM. 
The classifier output is transmitted on the uplink by controlling the backscatter logic which modulates the load impedance of the backscatter node to change between reflective and non-reflective state. 

\mysub{Receiver.}
In addition to a backscatter node for energy harvesting and communication, our design uses a hydrophone (Omnidirectional Reson TC4014) that receives and decodes the FM0 encoded backscatter packets~\cite{h2a} . 

\subsection{Offline Pipeline}
For the evaluation, we used the Watkins Marine Mammal Sounds Database~\cite{data-watkins} and chose the sounds from eight mammals including Atlantic spotted dolphins, bearded seals, Beluga white whales, bottlenose dolphins, bowhead whales, harp seals, narwhals, and walruses. Overall we used 364 sound files and in each trial of our offline evaluation, we shuffled and split them as 80\%, 10\%, and 10\% for training, validating, and testing. We re-sampled each file with the sample rate (330) of our prototype's input device. The decoded sound signals were first normalized to $[-1.0,1.0]$ and then the first 512 signals were transformed into a spectrogram through STFT (window size = 64, window step = 32).

\begin{figure}[t]
    \centering
    \includegraphics[width=\linewidth]{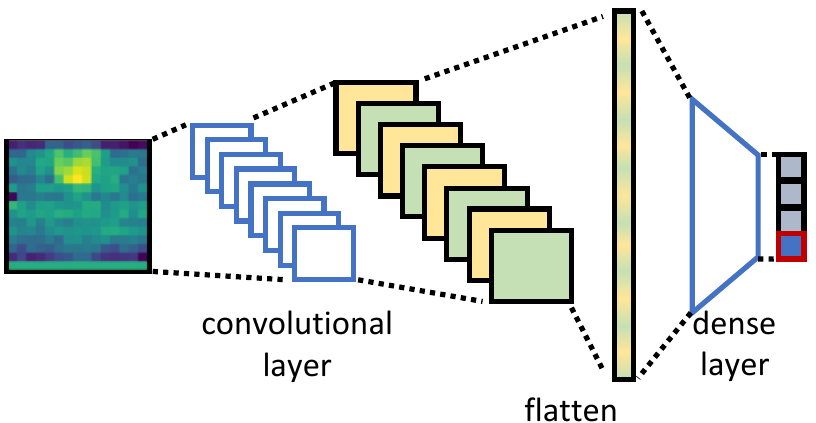}
    \caption{CNN structure for the classification task. \textnormal{The convolutional layer has 8 filters (size $3 \times 3$, strides $2 \times 2$) activated by ReLU. The dense layer is activated by softmax.}}
    \label{fig:cnn}
\end{figure}

We used TensorFlow to train high-level Keras models and then used Keras2C~\cite{conlin2021keras2c} to convert them into low-level models that are supported by standard C libraries. This library generates smaller DNN models than TensorFlow Lite does and requires fewer third-party libraries when generating models. We used a lightweight convolutional neural network (CNN) model as shown in Fig.~\ref{fig:cnn}. The spectrogram of normalized input signals is first reshaped as a 2-D input layer with size of $N \times N$. The input layer is connected to a convolutional layer that has 8 filters, each of which has the kernel size of $3 \times 3$ and strides of $2 \times 2$. The activated output of the convolutional layer is flattened and connected to a dense layer, which outputs the probability distribution of a classification.

\subsection{Model Accuracy}
\begin{figure}[!t]
    \centering
    \includegraphics[trim={0 0 0 1cm}, clip, width=\linewidth]{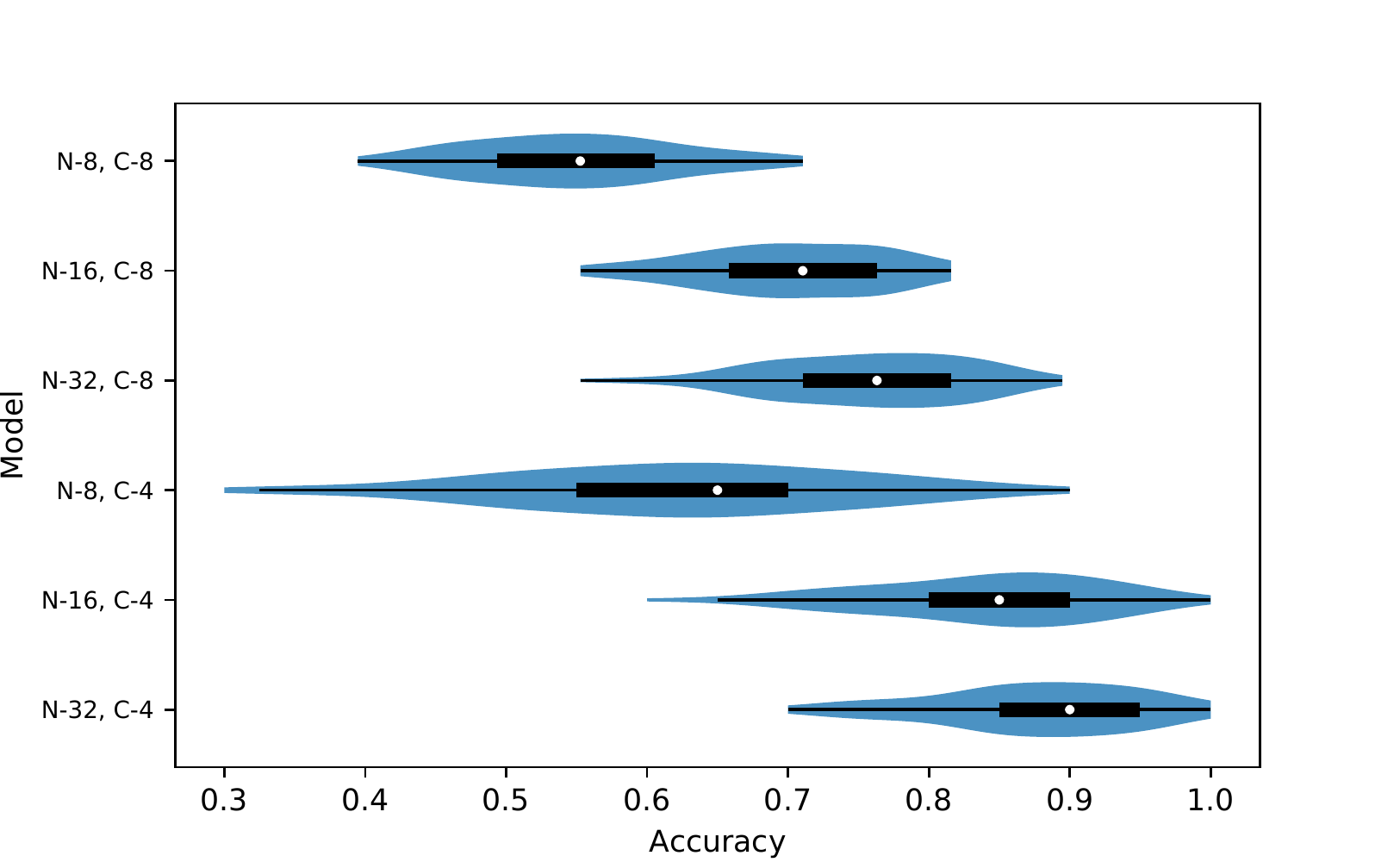}
    \caption{Distribution (boxplots and kernel density) of offline test accuracy CNN models with different input layer size (N) and number of mammal classes (C). \textnormal{The converted model of (N-16, C-4) can be deployed on our prototype.}}
    \label{fig:box_accuracy}
\end{figure}

We evaluated the offline accuracy of the model with different size $N$ of the input layer and number of mammal classes, as these two hyperparameters affected the size of the model. For 4-mammal classification, we used the data from Atlantic spotted dolphins, bearded seals, beluga white whales, and narwhals. Fig.~\ref{fig:box_accuracy} shows the offline accuracy distribution of each model hyperparameter configuration from 64 trials. For 4-mammal classification, the converted models when $N=8$ and $N=16$ can fit in our prototype, while $N=32$ demands a larger model which didn't fit in the memory of our microcontroller. The average accuracy for $N=16$ and $N=32$ is 84\% and 88\% respectively. For 8-mammal classification, with $N=32$, the average accuracy is 76\%, which requires larger memory than that on our prototype.

We further evaluated the model on the online data generated from the ADC on our prototype. To this end, we deployed the converted 4-mammal classification model into the pipeline on our prototype. We then randomly selected 16 raw sound files (4 samples for each mammal class) from the dataset. We used SIGLENT SDG 1032X function generator~\cite{funcgen} to convert raw sound files (animal sounds~\cite{data-watkins} of whales, seals, dolphins, and narwhals) to analog signals that were fed to the ADC. For each input signal, the microcontroller sampled the input analogue signal at 330 samples/s, normalized it, and computed the spectrogram of the processed data before running inference on it. The average accuracy of the online model was 63\% (while random guess is 25\%) which is lower than the offline accuracy. The lower online accuracy can be attributed to low number of testing samples and loss of data resolution during sampling. Nonetheless, this result demonstrates that our prototype was successful in on-board classification of four marine animals with a reasonable accuracy at extremely low power, and this accuracy may be improved as the research evolves and with more comprehensive online evaluation.

\subsection{Power Performance}
Next, we used the MSP-EXP430FR5994 development kit board to evaluate the power consumption of the microcontroller while it was running the end-to-end pipeline. We computed the power as the product of the current and the voltage when the board was connected to a power supply. We used SIGLENT SDG 1032X function generator~\cite{funcgen} to power on the board at 1.9 V. To measure the current, we connected Nordic Semiconductor’s Power Profiler kit~\cite{powerprof} as an ammeter in series with the board. Table \ref{table:1} shows the power consumption of each stage of the pipeline. At 1.9 V input voltage, the microcontroller consumed 932 $\mu$W during ADC sampling, 1.3~mW while running inference on the data, and 902 $\mu$W when it was backscattering the inference result. These power measurements are in line with those expected from the microcontroller's datasheet. Since prior work has demonstrated the potential to harvest up to few milliwatts from underwater acoustic and ultrasonic signals~\cite{underwaterharvesting}, these results demonstrate that it would be possible for our node to operate entirely based on harvested energy.
\looseness=-1

\begin{table}[h!]
\centering
\begin{tabular}{ |p{2cm}||p{1.8cm}|p{1.2cm}|p{1.8cm}|  }
 \hline
 \textbf{Stage} & \textbf{Power ($\mu$W)} & \textbf{Time (s)} & \textbf{Energy (mJ)} \\
 \hline
 ADC Sampling   & 932 & 1.6 &   1.49\\
 Inference & 1300  & 3.0 & 3.91\\
 Backscatter & 902  & 0.012 & 0.0108\\
 \hline
 \multicolumn{4}{|c|}{\textbf{Total Energy Consumption: 5.40~mJ}} \\
\hline
\end{tabular}
\vspace{0.25in}
\caption{On-board power and energy analysis of the battery-free inference prototype with a data rate of 1 kbps and a sampling rate of 330 samples/s. }
\label{table:1}
\end{table}

\vspace{-0.1in}

Since the power budget for ADC sampling and backscatter communication is similar, one could argue that it may be more efficient to send the entire data to the base station rather than performing the on-board inference. However, recall from~\xref{sec:inference} that this approach will be highly inefficient because of the limited throughput of the underwater channel. To demonstrate this inefficiency, consider the following example: For a backscatter node transmitting at 1 kbps (which is standard throughput for underwater acoustic modems), it will take around 6.14 seconds to send 512 samples of raw data (total \# of bits = 512$\times$12). The ADC consumes 1.49~mJ of energy to sample 512 samples at a sampling rate of 330 samples/s. This translates to 7.03~mJ of required energy. On the other hand, performing on-board inference, which typically takes 3 seconds (consuming 3.91~mJ) then transmitting the 12-bit-long inference result (consuming 0.0108~mJ) would require a total of 5.40~mJ, which makes sending raw data 30.19\% more power hungry than sending inference data.
This result demonstrates that battery-free edge inference is feasible, and that it makes on-board inference both more efficient and faster, even with this preliminary inference model. We envision that the gains can be significantly improved as the research evolves (see~\xref{sec:discusstions}).
\looseness=-1

%% file: 4_related.tex
\section{Related Work}
\label{sec:related}

\mysub{Battery-free underwater sensing.}
The past few years have seen multiple advances in underwater battery-free sensing and computing~\cite{jang2019underwater,guida2020underwater,ghaffarivardavagh2020ultra, hu2021sparse}. These advances have been propelled by three key trends. First is the downward trend in the power consumption of electronics, which has resulted in ultra-low-power processors (micro-controllers and FPGAs) that can operate in the micro-Watt regime~\cite{MSP,nanoigloo}. The second trend is the development of ultra-low-power underwater communication mechanisms, specifically piezo-acoustic backscatter~\cite{jang2019underwater,ghaffarivardavagh2020ultra}, which operates at net-zero power. Third is the continued improvement in the efficiency of energy harvesting from ambient underwater sources (sound, movements~\cite{eelharvesting}, and temperature gradients~\cite{thermal}). The combination of these factors has led to underwater battery-free sensors that can compute and communicate while powering up entirely from harvested energy. Our work builds on this line of research and extends it via battery-free inference, paving the way for a new generation of underwater battery-free devices capable of learning and inference.

\mysub{Low-power machine learning.}
The past few years have also seen significant advances in low-power machine learning models. Researchers have developed techniques to reduce the size and power consumption of advanced learning models to make them suitable for running on mobile and edge devices. 
These techniques include knowledge distillation, model pruning, quantization, or replacing a model's structure with a suitable one through network architecture searching~\cite{hinton2015distilling,elsken2019neural,mo2020darknetz}. Motivated by these advances, multiple research projects have explored the potential to perform machine learning and inference on battery-free devices~\cite{lee2019intermittent,montanari2020eperceptive,nardello2019camaroptera,gobieski2019intelligence}. However, none of these past systems are suitable for underwater environments since they rely on RF and/or large solar panels to harvest energy - neither of which are available in the deep sea. Our research is motivated by a similar desire to enable battery-free machine learning and inference, and is the first to bring such capabilities to underwater environments.

\mysub{Machine learning for underwater applications.} Finally, there has been significant interest in applying machine learning for various underwater problems, such as recognition of marine animals using their sound or images~\cite{zhuang2017marine, mellinger2006mobysound}, classifying seabeds~\cite{frederick2020seabed}, robotic navigation~\cite{gunnarson2021robotic}, and image enhancement~\cite{yang2021laffnet}. However, past systems for underwater ML required instruments with dedicated energy sources (typically underwater drones). Our research builds on this line of work and is the first to demonstrate the potential for battery-free underwater inference and machine learning.

%% file: 5_discussions.tex
\section{Open Problems and Opportunities}
\label{sec:discusstions}
In this paper, we proposed and prototyped battery-free ML inference in underwater environments as an exciting challenge for the community. We have shown that a lightweight CNN model can achieve decent accuracy while its on-board energy requirements can be supported by these devices. Our initial results indicate feasibility of performing sensing and inference. 
Future potential directions include:

\mysub{Battery-free on-device personalization.} 
Our design assumed that the same low-level model would be deployed on all battery-free devices deployed underwater. However, 
when implementing battery-free ML at a large scale, we expect that the sensed data by different underwater devices may differ at different locations (\eg, animal sounds near or far from a coast) and with environmental factors (\eg, temperature, pressure, multipath all impact sounds). Therefore, models on the devices would need to be locally tuned to fit different testing data distributions. Previous research~\cite{mo2021ppfl} proposes to use early exit during offline training so that an end-device 
can personalize a part of the model to improve its testing accuracy. To enable such on-device personalization, light-weight model training algorithms that can satisfy the computational constrains on battery-free devices are needed.

\mysub{Model-optimized hardware/software design.}
Currently there are limited compilers that can build DNN models on ultra-low power MCUs (\eg, the MSP430 class) such as~\cite{sliwa2020limits}. Co-design of task-specific hardware components alongside the ability to compile advanced models for lower-power MCUs can enable a wealth of new environmental monitoring applications. Specifically, one direct approach is to adopt existing optimization techniques for online inference on a target platform,
such as post-training quantization~\cite{liu2021post,nagel2020up}. It can reduce the sizes of on-board models and increase computational efficiency by avoiding floating operations. Another ``engineering'' solution is to optimize the implementations of online inference for different target platforms by using hardware-specific instructions and libraries that offer faster or more energy-efficient operations. An auto-design pipeline for such individualization would be worthwhile considering the diversity of IoT devices and used low-power MCUs.


\mysub{Battery-free distributed ML training.}
Given the potentially large scale deployment of low-cost and low-pollution battery-free nodes in the ocean, another future direction is to continuously sense data and train ML models in a decentralized manner among participating nodes. This will enable life-long ML with up-to-date data that reflects the changing nature of underwater environments, which is critical for many application domains including climate change and biodiversity monitoring. Similar to on-device personalization, the key challenge is to have light-weight ML training algorithms on the devices. Another challenge is the difficulty of obtaining labelled data from underwater environments. One option is to use unsupervised ML to only train feature extractors (\eg, autoencoders) that can help train classifiers on a cloud server~\cite{Zhao2021semisupervised}; another option is to generate pseudo labels on local data through data augmentation~\cite{jeong2021federated}. In addition, distributed/federated ML solutions that can coordinate devices, stations, and cloud servers are also needed to manage the local training tasks, to make sure that they can satisfy the energy constraints of battery-free underwater devices.

\vspace{0.1in}

\noindent
\textbf{Acknowledgments.} We thank our shepherd, the anonymous reviewers, and the Signal Kinetics group for their feedback. This research is supported by the Office of Naval Research, the MIT Media Lab, the Doherty Chair in Ocean Utilization, and the MIT-Imperial Seed Fund. 

\vspace{0.2in}